\documentclass{sig-alternate-05-2015}
\usepackage{acronym}
\usepackage{subfig}
\usepackage{booktabs}
\usepackage{array}

\acrodef{d2d}[d2d]{Document to Document}
\acrodef{s2s}[s2s]{Sentence to Sentence}
\acrodef{ts2ts}[ts2ts]{Topic Segment to Topic Segment}
\acrodef{d2ts}[d2ts]{Document to Topic Segment}
\acrodef{d2w}[d2w]{Document to Word}
\acrodef{d2s}[d2s]{Document to Sentence}
\acrodef{MOOC}[MOOC]{Massive Open Online Course}
\acrodef{POS}[POS]{Part of Speech}
\acrodef{AVL}[AVL]{Adelson-Velsky and Landis'}
\acrodef{4S}[4S]{Similar Segments in Social Speech}
\acrodef{IR}[IR]{Information Retrieval}
\acrodef{CST}[CST]{Cross-document Structure Theory}
\acrodef{ASR}[ASR]{Automatic Speech Recognition}
\acrodef{DBSCAN}[DBSCAN]{Density-Based Spatial Clustering of Applications with Noise}
\acrodef{KDE}[KDE]{Kernel Density Estimation}
\acrodef{NMF}[NMF]{Non-negative Matrix Factorization}
\acrodef{CNM}[CNM]{Clauset-Newman-Moore}
\acrodef{RBF}[RBF]{Radial Basis Function}
\acrodef{BST}[BST]{Binary Search Tree}

\newcolumntype{C}[1]{>{\centering\arraybackslash}p{#1cm}}
\newcolumntype{R}[1]{>{\raggedleft\arraybackslash}p{#1cm}}
\newcommand{\specialcell}[2][c]{%
	\begin{tabular}[#1]{@{}c@{}}#2\end{tabular}}

\begin{document}
\setcopyright{acmcopyright}

\doi{XX.XXX/XXX_X}

\isbn{XXX-XXXX-XX-XXX/XX/XX}

\conferenceinfo{CIKM '16}{October 24--28, 2016, Indianapolis, IN, USA}

%

\title{Graph-Community Detection for Cross-Document Topic Segment Relationship Identification}
%
%
%
%
%

\numberofauthors{3} 
%
\author{
%
%
\alignauthor
Pedro Mota\\
\affaddr{Instituto Superior T\'{e}cnico\\
Lisboa, Portugal\\
Carnegie Mellon University\\
Pittsburgh, PA, USA}\\
\email{pedrom@andrew.cmu.edu}
\alignauthor
Maxine Eskenazi\\
\affaddr{Language Technologies Institute\\
	Carnegie Mellon University\\
	Pittsburgh, USA}\\
\email{max@cs.cmu.edu}
\alignauthor
Lu\'{i}sa Coheur\\
\affaddr{Instituto Superior T\'{e}cnico\\
	Lisboa, Portugal}\\
\email{luisa.coheur@inesc-id.pt}
}


\maketitle
\begin{abstract}
In this paper we propose a graph-community detection approach to identify cross-document relationships at the topic segment level. Given a set of related documents, we automatically find these relationships by clustering segments with similar content (topics). In this context, we study how different weighting mechanisms influence the discovery of word communities that relate to the different topics found in the documents. Finally, we test different mapping functions to assign topic segments to word communities, determining which topic segments are considered equivalent.

By performing this task it is possible to enable efficient multi-document browsing, since when a user finds relevant content in one document we can provide access to similar topics in other documents. We deploy our approach in two different scenarios. One is an educational scenario where equivalence relationships between learning materials need to be found. The other consists of a series of dialogs in a social context where students discuss commonplace topics. Results show that our proposed approach better discovered equivalence relationships in learning material documents and obtained close results in the social speech domain, where the best performing approach was a clustering technique.
\end{abstract}


%
%


%
%

%
%



\section{Introduction}
Exploring a document in order to find specific content can be time consuming~\cite{Malioutov06,Sil11}. For example, skimming a video to determine where a particular topic is addressed is cumbersome since the video might have a considerable length. If only the raw video is provided, the only option is to randomly skim it, which is inconvenient and might lead to important parts being skipped. This problem is further aggravated when multiple documents from different sources and different media need to be considered. Identifying cross-document relationships addresses this problem. The relationships allow a more efficient browsing of the documents since as soon as we are able to find relevant information on one document, pointers to related content in other documents can be provided. In this context, we focus on equivalence relationships between documents. The nature of the relationships is defined from a user perspective rather than a strict semantic equivalence. This means that when users find some specific information in a document they might request ``more like this'', which does not mean that semantically equivalent content needs to be provided. An example of an application scenario for this situation is when students browse different learning materials looking for a specific part of a lecture they did not understand. Since different students might learn more efficiently with different learning materials, it is useful to have quick access to equivalent topic segments.

\begin{figure}[!ht]
	\centering
	\includegraphics[scale=0.23]{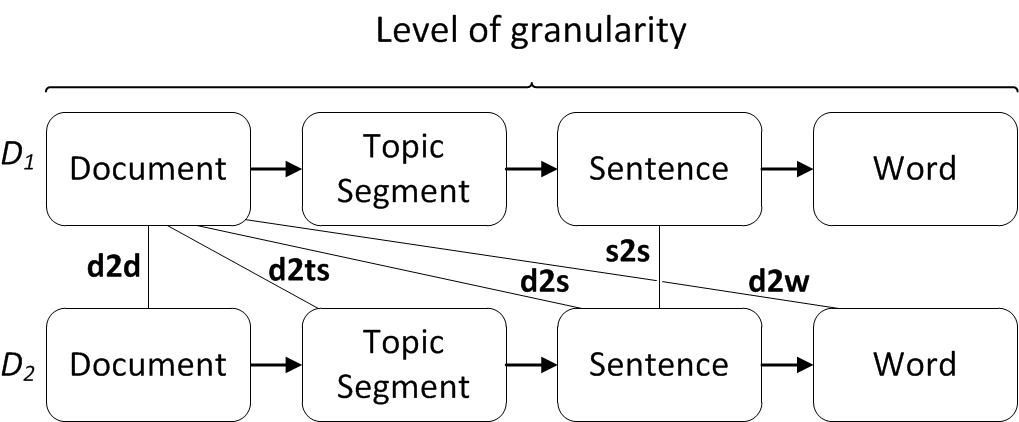}
	\caption{Taxonomy of cross-document relationships.}
	\label{fig:relsTax}
\end{figure}

The access to information in different documents can be done efficiently if a document's content structure can be perceived~\cite{Malioutov06, Allan98}. In the literature we find distinct ways of highlighting the structure of documents based on establishing relationships between them. The main difference in the approaches is the level of granularity of the relationships. A schematic representation of the types of relationships found in the literature is in Figure~\ref{fig:relsTax}, where we depict the different constituents of documents: words, sentences, topic segments, and the document itself. Depending on the constituents, different relationships can be established, namely: \ac{s2s}, \ac{d2w}, \ac{d2s}, \ac{d2ts}, and \ac{d2d} relationships.

The level of granularity of the relationships determines the types of support that are provided. For instance, when a word in a document is linked to another document (\ac{d2w}), the user expects to find information that can help him better understand the concepts regarding that word, whereas when two documents are linked (\ac{d2d}), users expect that these share a common general topic. This is helpful when in-depth knowledge about a topic is necessary. Despite the usefulness of these relationships, they cannot help when users want to access information regarding specific topics discussed in the documents. To deal with this, cross-document relationships need to be established at the topic segment level~(\ac{ts2ts} relationships). Other relationships will require users to manually skim the documents. By using \ac{ts2ts} relationships it is possible to have immediate access to content on the same topic in different documents. Current research work has given proof that structuring documents based on \ac{d2d} relationships brings both browsing efficiency and understanding of the topics in the documents~\cite{Shahaf12,shen15}. Therefore, we believe that similar advantages should also be brought to document topic segments. To the best of our knowledge, the problem of automatically finding cross-document relationships in topic segments is novel, as previous research work as focused on other levels of granularity. 

\begin{figure}[!ht]
	\centering
	\newcommand{\imgwidth}{0.42}
	\subfloat[Subfigure 1 list of figures text][Example of a \ac{d2w} relationship.]
	{
		\includegraphics[width=\imgwidth\textwidth]{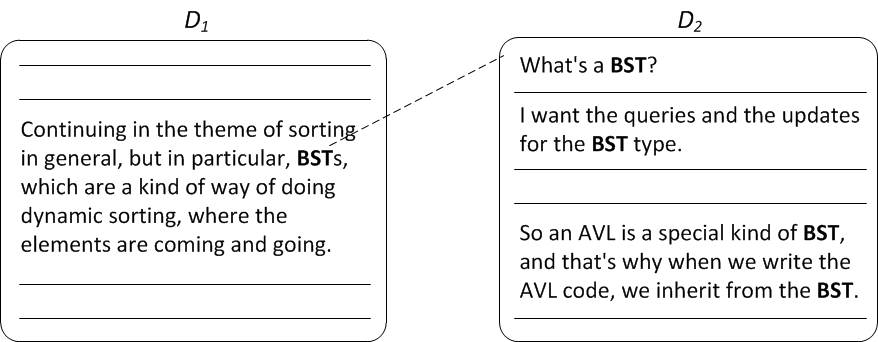}
		\label{fig:d2w}
	}\\
	\subfloat[Subfigure 2 list of figures text][Example of a \ac{ts2ts} relationship.]
	{
		\includegraphics[width=\imgwidth\textwidth]{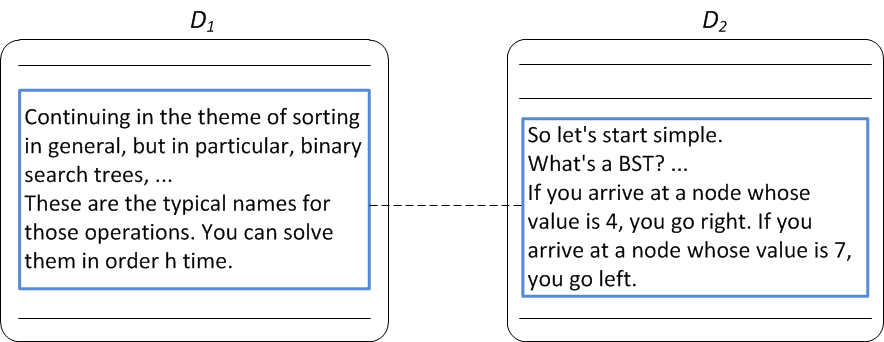}
		\label{fig:ts2ts}
	}
	\caption{Cross-document relationships in the \ac{AVL} domain.}
	\label{fig:cd-rels}
\end{figure}

To motivate the use of \ac{ts2ts} relationships for browsing documents, Figure~\ref{fig:cd-rels} provides an example that compares them to \ac{d2w} relationships. The example is in an educational domain on \ac{AVL} trees, a subject taught in Computer Science curricula. The example corresponds to manual transcripts from video lectures. In Figure~\ref{fig:d2w}, the target word is ``BST'' (Binary Search Tree). From this example it is possible to understand why \ac{d2w} relationships are not helpful when specific information in a document needs to be found. In $D_2$ there are multiple occurrences of the word ``BST'', indicating that it is used in different topics of the lecture. The problem is that from the word we can only reach a document, which gives no indication of the context where it was used. Even if the word is highlighted, such as in the example, we still would not know which of the occurrences is the appropriate one. This contrasts with the \ac{ts2ts} relationship in Figure~\ref{fig:ts2ts}, where it is possible to easily find in both documents where the topic of binary search trees is covered. Since the relationship is established at the topic segment level on both ends, users can immediately analyze the corresponding text, since it is topically self-contained.

The research area in which this work is framed is graph-community detection~\cite{Reichardt06, Yang13}. We propose the use of these methods to explore a co-occurrence graph structure, which models how words relate in different documents. The goal is to use the discovered word communities to perform cross-document relationship identification. We hypothesize that approaches based on text similarity, such as clustering, cannot cope with this task, since a substantial part of the vocabulary is shared in related documents. In this context, one of the challenges consists of studying ways to use the discovered words communities to identify cross-document topic segment equivalence relationships.

Our main contributions are:
\begin{itemize}
	\item Create an annotated corpus for studying topic segment relationships.
	\item Formalize a graph-community detection technique to automatically discover topic segment relationships.
	\item Evaluate the task of topic segment relationship identification with techniques from different research areas.
\end{itemize}

\section{Related Work}
\label{sec:rw}
Different approaches to document relationship identification can be found in the literature. As previously mentioned, what characterizes them is the level of granularity at the which the relationships are established (words, sentences, topic segments, or documents). Based on these levels of granularity, this section summarizes the research done in this area.

The \textit{Wikify!} system~\cite{Mihalcea07} identifies relationships between keywords in a document and Wikipedia articles (\ac{d2w} relationships). The difficulty in this task is the ambiguity of the extracted keyword. For example, if the keyword ``bar'' is extracted from a document about music, then it is necessary to relate it to an article with the same context. To address this problem a classification approach was used.

The identification of \ac{d2d} relationships has been performed in different scenarios. For example, the work described in~\cite{shen15} finds \ac{d2d} relationships with the goal of structuring lectures from \acp{MOOC} into a graph. The motivation is that the way of searching in \ac{MOOC} platforms is to enter a query and examine the retrieved list of courses. The problem is that a single course might not have all the information. To address this issue, prerequisite and equivalence relationships are established between lectures.

Another example where \ac{d2d} relationships are established is described in~\cite{Shahaf12}. In this work the problem of structuring a set of related news articles is addressed. The goal is to highlight how different documents contribute to the main overall topic. The structuring of the documents is based on the metro map metaphor. A metro map of documents consists of a set of lines, which can intersect each other. Each individual line follows a coherent narrative thread, covering a specific aspect of the main topic being analyzed. In turn, each metro line is composed of a set of metro stops. These metro stops correspond to clusters of words. Associated to each metro stop are the corresponding documents that originated them. The line intersections of their lines show that different aspects of a topic have some sort of relation. The generation of metro maps is based on the bigclam graph-community detection algorithm~\cite{Yang13}, which analyses a word co-occurrence graph in order to identify word communities.

The use of \ac{s2s} relationships is studied in \cite{Radev04, Maziero10}. This work uses a linguistic taxonomy for describing the different types of sentence relationships. The taxonomy is defined within the context of the \ac{CST}~\cite{Radev00}, whose goal is to understand how different parts of different documents related to one another. The \ac{CST} describes a total of 24 domain-independent semantic relations, such as ``Elaboration'', ``Contradiction'', or ``Equivalence''. This contrasts with previous work where either no specific relationship was attributed or a more addhoc approach was used. The approach to automatically identify \ac{s2s} relationships used a classification approach.

In what respects \ac{d2ts} relationships, the work described in \cite{Kokkodis14} uses them to link sections of a textbook and video lectures. To discover the relationships, the first step consists of constructing a set of queries based on keywords extracted from the textbook. The queries are then submitted to a video search engine. The task of identifying \ac{d2ts} relationships is viewed as an optimization problem where the minimal set of textbook sections to cover a video needs to be found.

The \ac{4S} task in the MediaEval campaign~\cite{SSSS} is the most closely related to ours, since it targets \ac{ts2ts} relationships. The task is defined as receiving an audio segment of interest and returning an ordered list of jump-in points for regions similar to it. By doing so, whenever a relevant piece of information is found, other related information can also be accessed efficiently. For this purpose, a corpus of dialogs among university students was created. Some examples of the topics discussed in the dialogs are: classes, new technologies, or movies. An example of a work that addresses this task is described in~\cite{Galu14}. This study assesses the influence of Passage Retrieval when approaching the task from an \ac{IR} perspective. This is done by creating supervised segmentation models.

The main difference between our work and the \ac{4S} task is that of social speech context. Dialogs in this context do not follow an overarching topic. This means that they contain a variety of topics which are not related to each other. In this work we propose the study of topic segment relationships given a set of related documents. Therefore, we expect the topic segments to be much more similar, due to the interweaving of their related topics.

\section{Identifying Topic Segment Relationships}
\label{sec:rel_id}
In this section, we describe our graph-community detection-based approach to the \ac{ts2ts} relationship identification task. Given that we want to group similar document topic segments, we can formalize this problem in a clustering setting:
\begin{description}
	\item[Input:] A set of items $S_1, ..., S_m$. Each item corresponds to the textual content of a topic segment. 
	
	\item[Output:] A mapping from each item to a particular cluster $k \in 1, ..., K$.
\end{description}

To perform the clustering task we propose the use of graph-community detection techniques. The idea is to analyze a weighted co-occurrence graph representation of document segments and find word communities that are representative of the topics in the documents. This is similar to the Metro Maps approach in \cite{Shahaf12}. We hypothesize that co-occurrence graphs can better model how words relate to one another in different documents. This contrasts with cluster approaches where features relate to individual words and the presence or absence of the features is based on whether the corresponding word occurred in the document or not.

The graph-community detection problem can be formalized as follows:
\begin{description}
	\item[Input:] a weighted co-occurrence graph $G_{co} = (W, E)$, where $W$ is the set of nodes and $E$ the set of edges. $W$ corresponds to the set of words from a given set $S$ of document segments. An edge $(w_i, w_j)$ exists if words $w_i$ and $w_j$ occur in some segment $S_i \in S$.
	
	\item[Output:] a mapping from each word $w_i \in W$ to a particular community $c \in 1, ..., C$.
\end{description}

Appropriately setting the weights $w(i,j)$ of the edge is a topic of research in this work. Depending on how these weights are set, different word communities can be obtained. Therefore, it is necessary to develop an appropriate weighting scheme to the cross-document topic relationship identification task. In a related problem described in \cite{Shahaf12}, a weighting scheme that counts the number of co-occurrences between words is used to discover a Metro Map structure of news articles. We hypothesize that having all word co-occurrence contribute in the same way to a weight score might not work in the document segment relationship task. Therefore, we designed the following \textit{tf-idf}-based weighting schemes:
\begin{itemize}
	\item \textbf{Count}: the number of times the words co-occurred in different segments (equivalent to \cite{Shahaf12}).
	\item \textbf{Best \textit{tf-idf}}: the sum of the highest \textit{tf-idf} values of the words.
	\item \textbf{Count + Best \textit{tf-idf}}: the sum of the previous weights.
	\item \textbf{Count + Avg \textit{tf-idf}}: the sum of the \textit{count} weight and the sum of the average \textit{tf-idf} values of the words.
\end{itemize}

After obtaining the word-communities, it is necessary to perform a mapping between them and the topics segments. This task needs to be performed since the previous step only provides groups of words. Therefore, it is necessary to determine how to group the topic segments. Doing this using word communities is, to the best of our knowledge, novel. It should be noted that although the work in \cite{Shahaf12} used graph-community detection, the problem being solved is different. The relationships between the documents were given \textit{a priori} by grouping documents if they occurred in the same time span. The graph-community detection task was then carried out in those groups of documents to determine the metro stops. Our work is different since we cannot rely on this criterion to discover the document relationships. In this work we propose the use of a function to assign scores between a document segment and word communities. Segments are assigned to the highest scoring community. If two document segments are assigned to the same community they are considered as equivalent. The mathematical formula that expresses this idea is defined as follows:
\begin{align}
\underset{c \in C}{argmax} \mbox{ } score(seg, c),
\end{align}
where $C$ is the set of communities discovered in $G_{co}$, and $seg$ is the set of words in a segment. It should be noted that different formulations of the $score$ function can be designed. In the experiments to identify similar document segments we considered the following scoring functions:
\begin{align}
score_c(seg, c) = \frac{|seg \cap c|}{|c|},\label{eq:sf1} \\
score_{seg}(seg, c) = \frac{|seg \cap c|}{|seg|}, \label{eq:sf2} \\
score_{tfif}(seg, c) = \frac{\displaystyle\sum_{w_i}^{|seg \cap c|} tfidf(w_i)}{\displaystyle\sum_{w_i}^{seg} tfidf(w_i)}  \label{eq:sf3}
\end{align}

The first two functions count the common words between the segment and the community. The score is then normalized either by the total number of words in $c$ or $seg$, respectively. The previous functions treat all words in the same way. Therefore, we also defined a function that makes words contribute according to their relevance, $score_{tfidf}$. The difference is that common words have a score corresponding to their normalized \textit{tf-idf} value.

\subsection{Evaluation}
In this section, a description of the evaluation procedures for the cross-document \ac{ts2ts} relationship identification task is provided. For the evaluation a comparison between techniques from different research areas was carried out. The following research areas were surveyed: clustering, topic models, and graph-community detection. The evaluation was carried out in two different domains: a corpus with \ac{AVL} trees learning materials, and the \ac{4S} task~\cite{SSSS}.

\subsubsection{Measures}
\label{sec:clus_metric}
Since the cross-document relationship identification task is viewed as a clustering problem, standard evaluation metrics of this area were used. One of these metrics is the Adjusted Rand Index (ARI)~\cite{Vinh09}, which computes the similarity between clusterings by analyzing all pairwise combinations of data points in a predicted clustering and a ground truth clustering.

Another standard cluster quality measure is the $F_1$ score. In clustering precision and recall are computed over pairs of items. For example, a true positive means that both the hypothesized clustering and the true clustering assigned some pair of items to the same cluster.

The last evaluation metric we consider is $Accuracy$, which measures the percentage of items that are assigned to their correct clusters. This requires a mapping between the hypothesized clusters and the true clusters. This mapping is found using the Kuhn-Munkres algorithm~\cite{Kuhn55}.

\subsubsection{Corpora}
\label{sec:relid_corp}
The evaluation of the cross-document relationship identification task presents the problem of not having available datasets with topic segment annotations. The only exception is the \ac{4S} corpus~\cite{SSSS}. The problem with this corpus is that it contains social spontaneous speech. Therefore, there is not an overall topic, since the dialogs have unrelated topics. Another characteristic of the corpus is its user-centered approach to the annotation. This means that the use case for the annotation is a scenario where users are browsing a set of related documents and when relevant information is found they request ``more like this''. This means that users' needs have priority over annotating semantic equivalence relationships in a strict way. Despite not being ideal, we still consider it in our evaluation. The \ac{4S} dataset contains a total of 25 dialogs, with a total duration of approximately 4 hours, 26 different topics, and a total of 185 document segments. 

In order to use the \ac{4S} corpus it was first necessary to align the transcripts with the segment annotations. A perfect alignment could not be obtained because the time stamps in the annotations and in the transcripts did not have an exact match. Given these circumstances, we opted to consider as part of a segment any portion of the transcripts that overlapped with the segment annotations. In the evaluation only the manual transcripts were used, since the ones provided by an Automatic Speech Recognizer had a high error rate. Another problem with this corpus is that some of the topics allow a broad scope of segments to be grouped. For instance, the label ``Computer Science Topics'' contains segments discussing classes students are taking, assignments, career paths, \textit{etc}. Given this situation, we opted to choose a subset of the corpus containing more coherent topics. In this context, we used 8 topics with a total of 106 segments. A description of this subset is provided in Table~\ref{tab:corpus_ssss_ri}.
\begin{table}[!ht]
	\centering
	\begin{tabular}{C{1.7}C{1.5}C{1.5}C{1.3}}
		\toprule
		\textbf{Topic} & \textbf{Segments} & \textbf{\#Words} & \textbf{\#Vocab} \\
		\midrule
		\textbf{Courses} & 37 & 889 & 341\\
		\textbf{Math} & 7 & 209 & 110\\
		\textbf{Internships} & 6 & 69 & 46\\
		\textbf{Research} & 8 & 656 & 266\\
		\textbf{Family} & 7 & 208 & 130 \\
		\textbf{Games} & 12 & 1182 & 458\\
		\textbf{Movies} & 17 & 410 & 192\\
		\textbf{TV Shows} & 12 & 646 & 270\\
		\bottomrule
	\end{tabular}
	\caption{\ac{4S} corpus used in the experiments.}
	\label{tab:corpus_ssss_ri}
\end{table}

Facing the problem of not having available corpora for the evaluation, we annotated a corpus of \ac{AVL} tree-related learning materials. The learning materials come from different media, namely: slides, video lectures, and Wikipedia articles. A summary of the characteristics of this corpus is given in Table~\ref{tab:corpus}. All of the documents refer to the topic of \ac{AVL} trees except for one Wikipedia article which specifically addresses tree rotations (an essential operation in \ac{AVL} trees). The corpus was segmented into topically cohesive segments by the first author, who has taught AVL trees in an Algorithms and Data Structures course. For the segmentation of video lectures, manual audio transcripts were available. The main guideline for segment annotation was to identify places where a topic shift occurred such that disregarding this topic would make it harder to follow the higher-level structure of the document, as suggested in~\cite{Eisenstein08}. This emphasizes that documents should be segmented individually, instead of, for instance, trying to fit parts of documents in a predefined list of subtopics. In total 86 segmentation boundaries were annotated, from a corpus containing 3181 sentences.
\begin{table}[!ht]
	\centering
	\begin{tabular}{R{1.5}C{.7}C{1.3}C{1.9}C{1.2}}
		\toprule
		& \textbf{Docs} & \textbf{Segs ($\overline{x}$)} & \textbf{Words ($\overline{x}$)} & \textbf{\#Vocab} \\
		\midrule
		\textbf{Slides} & 5 & 7 $\pm$ 1.2 & 1402 $\pm$ 683.5 & 776 \\
		\textbf{Video} & 3 & 11 $\pm$ 5.6 & 6396 $\pm$ 307.2 & 1265 \\
		\textbf{Wikipedia} & 2 & 7 $\pm$ 1.4 & 1195 $\pm$ 221.3 & 536 \\
		\bottomrule
	\end{tabular}
	\caption{AVL corpus description.}
	\label{tab:corpus}
\end{table}

The segmented \ac{AVL} corpus was then annotated with equivalence relationships. This annotation was done in the same spirit as the \ac{4S} corpus, using a user-centered approach. The difficulty in this task is in the presence of multiple topics in a single segment. The annotation process consisted of going through each document individually. In the case of the first document, the segments were tagged with the topic they discussed. On subsequent documents, it was assessed if the segments should be assigned with an existed tag or if a new tag should be created. After the annotation process, a total of 15 different relationships were found in the set of 86 segments. The experiments only used the topics that appeared in the majority of the documents (49 segments). A description of this subset of segments is provided in Table~\ref{tab:corpus_ri}.
\begin{table}[!ht]
	\centering
	\begin{tabular}{C{2.6}C{1.5}C{1.3}C{1.3}}
		\toprule
		\textbf{Topic} & \textbf{Segments} & \textbf{\#Words} & \textbf{\#Vocab} \\
		\midrule
		\textbf{BST} & 7 & 1822 & 284\\
		\textbf{Tree Height} & 5 & 1800 & 338 \\
		\textbf{Tree Rotation} & 13 & 3762 & 538\\
		\textbf{Tree Balance} & 13 & 3670 & 483\\
		\textbf{AVL Rebalance} & 11 & 8142 & 700\\
		\bottomrule
	\end{tabular}
	\caption{AVL corpus used in the experiments.}
	\label{tab:corpus_ri}
\end{table}

\subsubsection{Algorithms}
\label{sec:ri_algs}
In this section, we describe the algorithms tested in the cross-document relationship identification task. Since we frame the task in a clustering setting, we survey algorithms designed specifically for this task, namely:

\textbf{k-means~\cite{Lloyd06}:} given a $k$ value that specifies the number of clusters to be obtained, the process is based on minimizing a criterion function measures the distance of data points and cluster centroids.

\textbf{Agglomerative clustering~\cite{Maimon2005}:} a hierarchical clustering that uses a bottom-up approach, considering in the beginning all points as individual clusters. The procedure consists of a series of iterations, and, in each of them, two clusters are merged. Therefore, in each step it is necessary to decide which clusters to merge, which is done by using a similarity measure and criterion function.

\textbf{DBSCAN~\cite{Sander98}:} based on 2 values, a radius ($Eps$) and minimum number of neighbors ($MinPts$). When a point in a \mbox{$Eps$-neighborhood} contains at least $MinPts$ it is defined as a \textit{core} point. When a core point is found, a cluster is formed with all its neighbors. Then, an expansion process takes place by checking if the neighbors are also core points. In the positive case, these are added to the cluster.

\textbf{Mean Shift~\cite{Cheng95}:} the points are regarded as a sample of a distribution. By applying a kernel to each point a probability surface is generated. Depending on the kernel bandwidth, the resulting probability surface will vary. This surface is characterized by its peaks, which corresponds to clusters. The clustering procedure is based on iteratively shifting each point uphill, until the nearest peak is reached.

\textbf{Spectral Clustering~\cite{Weiss99}:} uses an adjacency matrix of a similarity graph between the points to cluster. Then, the first $k$ eigenvectors of the Laplacian matrix are calculated. The final step consists of using the k-means algorithm to obtain a clustering from the eigenvectors.

\textbf{\ac{NMF} clustering~\cite{Xu03}:} based on the \ac{NMF} matrix factorization technique that factorizes a non-negative matrix $V$ into two other non-negative matrices. The goal of the factorization is to obtain $V \approx W H$. The factorization is found by using an optimization approach. The non-negative property of \ac{NMF} makes the reduced space easy to interpret. Each element $W_{ik}$ indicates the degree of association of point $i$ with cluster $k$. Therefore, it suffices to take the highest value of $W_i$ to find the corresponding cluster.

In Section~\ref{sec:rel_id} we proposed an approach to document relationship identification based on word communities. This can also be viewed as a topic modeling task~\cite{Blei12}, since its output are groups of words that relate to a topic. Therefore, we experiment with the use of topic models. In topic models documents are assumed to be observed from a generative process that contains hidden variables, which represent the topic structure of the document. Each topic is a distribution over words and documents are mixtures of those topics. Finally, words are viewed as draws from one of the topics. Using the previous concept, a probabilistic graphical model estimates the distribution of the hidden variables.

Taking into account that we propose the use of the graph-community detection approach, we survey different techniques that can perform this task, namely:

\textbf{Label Propagation (LP)}~\cite{Raghavan}: the algorithm starts with each node assigned to a different community. The nodes are then visited and assigned with the label which is more frequent among its neighbors. The algorithm terminates when there are no changes in the labels.

\textbf{\ac{CNM}}~\cite{CNM04}: based on the \textit{modularity} criterion. High modularity means that dense connections are observed in nodes from the same community and sparse connections are observed between nodes from different communities. The algorithm evaluates for each node the modularity gain from removing the node from its community and placing it in its neighbors. If the modularity increases, the node is reassigned. The process is repeated until no changes in the graph are observed.

\textbf{Louvain}~\cite{blondel08}: similar to the previous algorithm, with the difference that in each iteration nodes in the same community are merged into a single node and the corresponding edge weights added.

\textbf{Walktraps}~~\cite{Pons06}: the algorithm is based on \textit{random walks}. A random walk means that we start at a node, we pick a neighbor at random and move to it, then repeat the procedure. By repeating this procedure it is possible to compute statistics about the visited nodes. The statistics are summarized in a transition matrix, which expresses the probability of going from one node to another through a random walk of length $t$. Using the distance metric, the problem of graph-community detection is then approached as a clustering task.

\textbf{Leading Eigenvector}~\cite{Newman06}: the algorithm starts by having all nodes belonging to the same community. In each iteration, the graph is split into two communities in a way that a significant increase in modularity is obtained. The split is determined by evaluating the leading eigenvector of the modularity matrix.

\textbf{Bigclam}~\cite{Yang13}: based on the optimization of a likelihood community membership.

\subsection{Results with the AVL Corpus}\label{sec:avl_exps}
In this section, we report the document relationship identification experiments done with the \ac{AVL} corpus.

\subsubsection{Clustering}\label{sec:clus_avl_exps}
Each of the clustering algorithms described in Section~\ref{sec:ri_algs} has a particular set of parameters that can be tested. In this study we set those parameters in the following way:
\begin{itemize}
	\item k-means: the number $k$ of clusters was set to 5, the number of subtopics in the corpus.
	
	\item Agglomerative Clustering: the following similarity metrics where assessed: cosine, Euclidean, and Gaussian. For the Gaussian metric, variances ($\sigma^2$) within the range from 1 to 500 (step size 1) were tested. Different merge functions were assessed, namely: ward, complete, and average.
	
	\item DBSCAN: all combinations of $Eps$, within the range of 0.1 and 0.9 (step size 0.1), and $MinPts$, within the range of 1 and 14 (step size 1), were tested. All the previously mentioned similarity metrics were tested.
	
	\item Mean Shift: the \ac{RBF} kernel was used and tested with bandwidth values from 1 to 1000 (step size 1).
	
	\item Spectral Clustering: the number of desired clusters was set to 5. The same similarity metrics referred previously were used to test different similarity graphs.
	
	\item \ac{NMF} Clustering: the number of desired clusters was set to 5.
\end{itemize}

The implementation used for all clustering algorithms, except \ac{NMF}, was the one provided by~\cite{sklearn_api}. The \ac{NMF}-based clustering implementation was the one in~\cite{He14}.

Using the previous experimental setup, the best results obtained are described in Table~\ref{tab:clus_bl}. Results show that most of the techniques obtained low scores in all evaluation metrics. The only exceptions were the Agglomerative and Spectral clustering. It should be noted that not all evaluation metrics agree as to the best technique, since Spectral clustering had higher scores in $ARI$ and $F_1$, whereas Agglomerative clustering had the best $Acc$ score. Also, there is not a strict correlation between the evaluation metrics. For example, Mean Shift obtained the lowest $ARI$, but would be considered as one of the best performing techniques in $F_1$. In this context, we set as a baseline the Spectral clustering technique, since it was the best in two of evaluation metrics and still performed well on the third one.
\begin{table}[!ht]
	\centering
	\begin{tabular}{cC{1.2}C{1.2}C{0.9}}
		\toprule
		\textbf{Clustering Algorithm} & $\pmb{ARI}$ & $\pmb{F_1}$ & $\pmb{Acc}$\\
		\midrule
		k-means & 0.011 & 0.34 & 0.31\\ \midrule
		Agglomerative & 0.11 & 0.32 &  \textbf{0.52}\\ \midrule
		DBSCAN & 0.03 & 0.26 &  0.27\\ \midrule
		Mean Shift & 0.009 & 0.35 &  0.29\\ \midrule
		Spectral & \textbf{0.15} & \textbf{0.36} &  0.41\\ \midrule
		\ac{NMF} & 0.046 & 0.22 &  0.45\\
		\bottomrule
	\end{tabular}
	\caption{Results of the clustering algorithms in the \ac{AVL} corpus. Best results are in bold.}
	\label{tab:clus_bl}
\end{table}

\begin{figure}[!ht]
	\newcommand{\imgwidth}{0.15}
	\centering
	\subfloat[Subfigure 1 list of figures text][Cosine]
	{
		\includegraphics[width=\imgwidth\textwidth]{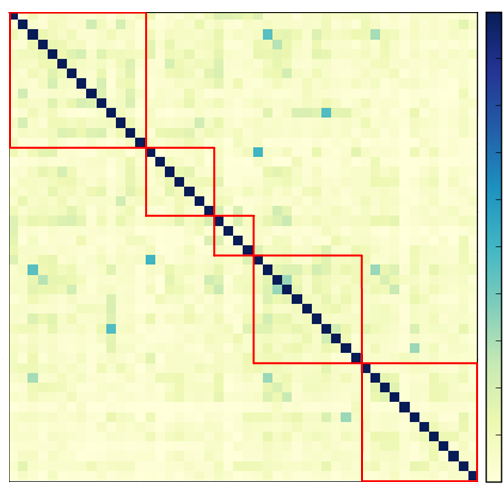}
		\label{fig:hm_cos}
	}
	\subfloat[Subfigure 2 list of figures text][Euclidean]
	{
		\includegraphics[width=\imgwidth\textwidth]{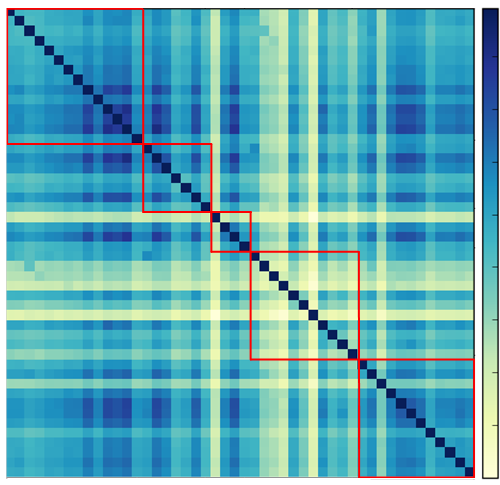}
		\label{fig:hm_eu}
	}
	\subfloat[Subfigure 2 list of figures text][Gussian]
	{
		\includegraphics[width=\imgwidth\textwidth]{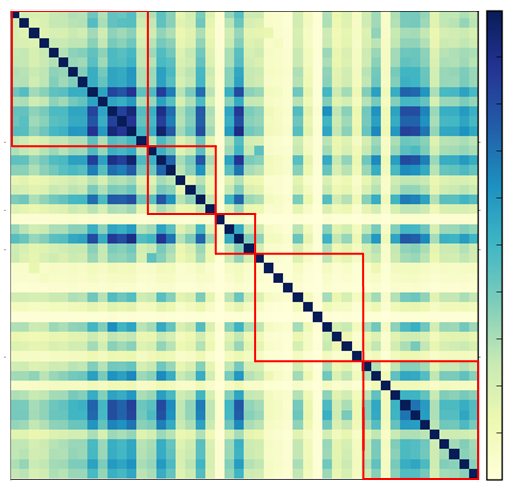}
		\label{fig:hm_gaus}
	}
	\caption{Heat maps of the \ac{AVL} corpus. The $C_i$ regions (delimited in red) correspond to the true clusters.}
	\label{fig:hms}
\end{figure}

In order to understand the space where the clustering task is performed, Figure~\ref{fig:hms} depicts the heat maps of the \ac{AVL} segment similarity matrix using the assessed similarity metrics. In the heat maps, darker tones indicate more similar segments. In all cases, one can observe that it is impossible to distinguish the true clusters. This means that no contrast can be found between segments that should belong to the same cluster and the ones that should not. The cosine similarity case (Figure~\ref{fig:hm_cos}) is the best example of this situation, since the map prominently has light colored cells. This means that all segments were considered not similar. The Euclidean and Gaussian cases (Figure~\ref{fig:hm_eu} and \ref{fig:hm_gaus}) present a better contrast. Despite this, it is still far from obvious what the clusters should be. In practice, most of the clustering techniques had a tendency to concentrate the majority of the segments in a single cluster. This constitutes a limitation that the similarity space imposes over the clustering techniques. This means that if the similarity space does not properly reflect how the segments should be grouped, then there is not much chance for the clustering techniques to perform well. This is a particularity of our task, since the segments are related to a common overarching topic and come from different media, which causes them to be wrongly perceived by a similarity metric.

\subsubsection{Topic Models}
The experiments in the context of topic modeling were made using software made available by \cite{rehurek10}. All the proposed scoring functions (Equations~\ref{eq:sf1}, \ref{eq:sf2}, and \ref{eq:sf3}) were tested. It was also necessary to choose the top-$n$ most probable words from each topic word distribution. For this purpose a range between the top-$1$ and top-$200$ words was assessed. The probabilistic nature of topic modeling makes it non-deterministic. This means that different runs will yield different topic distributions. Therefore, each of the topic models were run 10 times and the corresponding results averaged. 

The number of topics was set to 5, since it is the number of topics in the \ac{AVL} corpus. The results obtained were close between the tested scoring functions. Nevertheless, using $score_c$ with the top-$19$ words from each topic distribution obtains the best results: $ARI = 0.10$, $F_1 = 0.31$, $Acc = 0.44$, for $n = 19$. Therefore, this parameterization is our topic modeling baseline in the \ac{AVL} corpus.


\subsubsection{Graph-community Detection}
The experiments carried out in the \ac{AVL} corpus were performed by trying the graph-community detection techniques, described in Section~\ref{sec:ri_algs}, using all the proposed weighting schemes and scoring functions. It should be noted that some of the techniques are not sensitive to edge weight. The implementation of the techniques we use was the one provided by~\cite{igraph}. Another aspect to consider is that there are words in the documents that should not be taken into account, since they are either too common or too rare to be representative of a subtopic in a document. \cite{Shahaf12} addressed this problem using a \textit{tf-idf} to filter the vocabulary. In this experiment we also adopted this approach by setting for each segment a cutoff at the top-$100$ words with the highest \textit{tf-idf} score. A summary of the results is provided in Table~\ref{tab:gcd_bl}. It is possible to observe that the Walktraps algorithm performed better in all evaluation metrics.
\begin{table}[!ht]
	\centering
	\begin{tabular}{C{1.5}C{1.4}C{1.5}C{0.6}C{0.6}C{0.6}}
		\toprule
		\textbf{Algorithm} & \textbf{Weighting} & \textbf{Scoring Function} & $\pmb{ARI}$ & $\pmb{F_1}$ & $\pmb{Acc}$\\
		\midrule
		LP & - & $score_c$ & 0.0 & 0.35 & 0.27\\ \midrule
		\ac{CNM} & - & $score_{tfif}$ & 0.05 & 0.31 &  0.42\\ \midrule
		Louvain & Best \textit{tf-idf}& $score_c$ & 0.12 & 0.31 & 0.42\\ \midrule
		Walktraps & \specialcell{Count + \\ Avg \textit{tf-idf}} & $score_c$ & \textbf{0.19} & \textbf{0.39} & \textbf{0.48}\\ \midrule
		\specialcell{Leading \\ Eigenvector}& \specialcell{Count + \\ Best \textit{tf-idf} }& $score_{tfif}$ & 0.05 & 0.35 &  0.38\\ \midrule
		Bigclam & - & $score_{seg}$ & 0.06 & 0.1 &  0.25\\
		\bottomrule
	\end{tabular}
	\caption{Best results obtained with graph-community detection algorithms. In bold are the best results obtained.}
	\label{tab:gcd_bl}
\end{table}

The previous results indicate that it is important to differentiate the word co-occurrence weight according to an importance measure rather than just relying on word counting. As for the score function, the simpler $score_c$ function worked better. This demonstrates that after discovering the communities, words should be treated as equally important when performing the mapping task, otherwise higher \textit{tf-idf} words might induce unwanted bias. This makes sense, since if the words were assigned to the same community they represent some topic as a whole and, thus, a subset of the words should not be expressive enough to represent that topic.

\begin{figure}[!ht]
	\centering
	\includegraphics[scale=0.22]{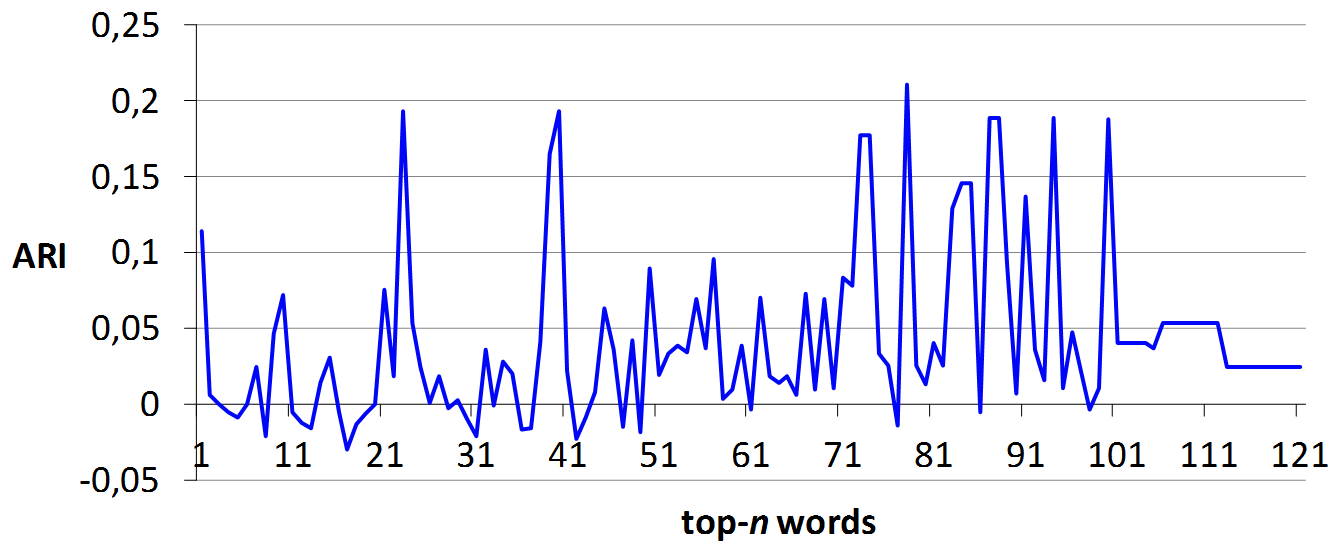}
	\caption{Results obtained by varying the top-$n$ words in Walktraps using Count + Avg \textit{tf-idf} and $score_c$.}
	\label{fig:walktraps_topn}
\end{figure}

Taking into account that exploring all possible top-$n$ words cutoffs is not feasible, since the number of parameter combinations is high, we only performed this analysis on the technique with the best previous results. In this context, we varied the number of words used between 1 and 300 and applied the Walktraps algorithm with Count + Avg \textit{tf-idf} and $score_c$. The results obtained are in Figure~\ref{fig:walktraps_topn}. The best result was obtained when $n = 78$ with  $ARI$, $F_1$, and $Acc$ values of 0.21, 0.44, 0.52, respectively. From the plot it is possible to observe that properly setting $n$ is important, since there is much oscillation in the results.

After the previous experiment, we used the best top-$n$ value and tested the length of the random walk parameter using values between 1 and 1000 (step size 1). The results show that there is no straight correlation between the increase of $t$ and the performance metric $ARI$. This makes it hard to set the optimal value of $t$. The best results were obtained for $t = 208$, with $ARI$, $F_1$, and $Acc$ values of 0.3, 0.46, 0.52, respectively.

\subsubsection{Discussion}
\label{sec:avl_res_dis}
Having performed all experiments with techniques from different research areas, this section provides a discussion on the overall results. In Table~\ref{tab:res_all} a summary of the best results is provided. The Walktraps approach provided the best performance in the document relationship identification task, obtaining better results in all evaluation metrics used. The most important conclusion from these results is that graph-community detection-based techniques are more suitable for the relationship identification task. The difference in the evaluation metrics is substantial, with a value twice as high in $ARI$, and a 10\% increase in $F_1$ and $Acc$. The problem with clustering techniques is that they rely on a similarity space, which is not appropriate for the nature of this task. Given the results obtained, we claim that the document relationship identification task should be done by exploring network properties of a co-occurrence graph using a graph-community detection approach.
\begin{table}[!ht]
	\centering
	\begin{tabular}{cC{2}C{1}C{1}}
		\toprule
		\textbf{Algorithm}& $\pmb{ARI}$ & $\pmb{F_1}$ & $\pmb{Acc}$\\
		\midrule
		Spectral Clustering & 0.15 & 0.36 & 0.41 \\ \midrule
		Topic Models & 0.10 & 0.31 & 0.44 \\ \midrule
		Walktraps & \textbf{0.30} & \textbf{0.46} & \textbf{0.52} \\
		\bottomrule
	\end{tabular}
	\caption{Summary of the results in the document relationship identification task.}
	\label{tab:res_all}
\end{table}

\begin{figure}[!ht]
	\centering
	\newcommand{\imgwidth}{0.25}
	\subfloat[Subfigure 1 list of figures text][True clusters.]
	{
		\includegraphics[scale=\imgwidth]{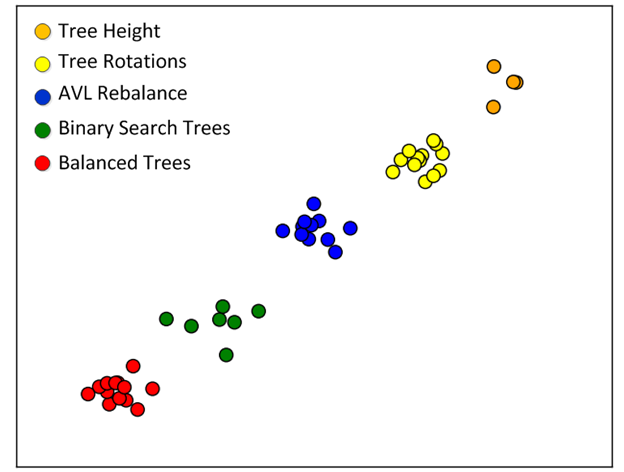}
		\label{fig:avl_true}
	}
	\subfloat[Subfigure 2 list of figures text][Spectral clustering.]
	{
		\includegraphics[scale=\imgwidth]{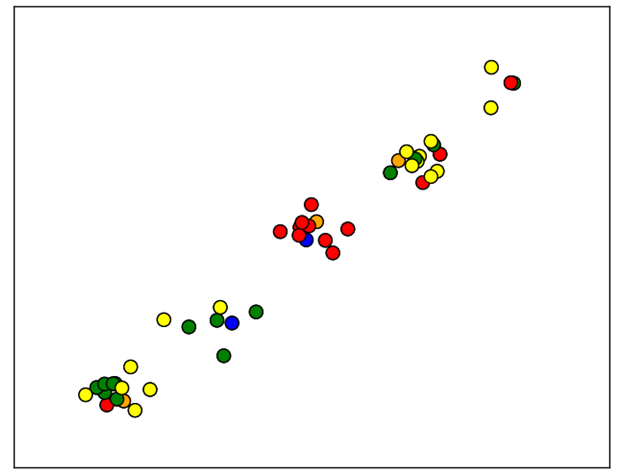}
		\label{fig:avl_spectral}
	}\\
	\subfloat[Subfigure 2 list of figures text][Topic Models clustering.]
	{
		\includegraphics[scale=\imgwidth]{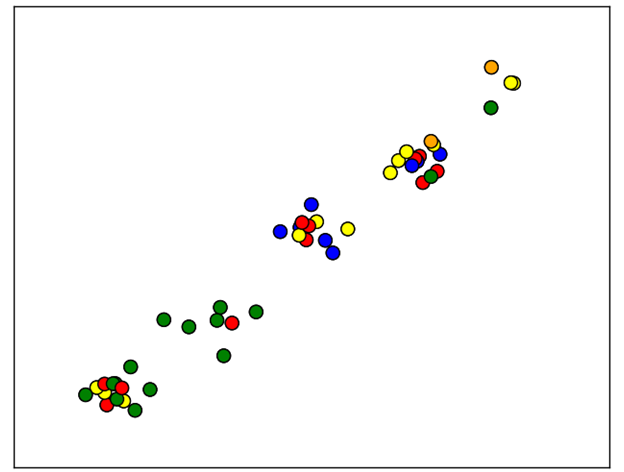}
		\label{fig:avl_lda}
	}
	\subfloat[Subfigure 2 list of figures text][Walktraps clustering.]
	{
		\includegraphics[scale=\imgwidth]{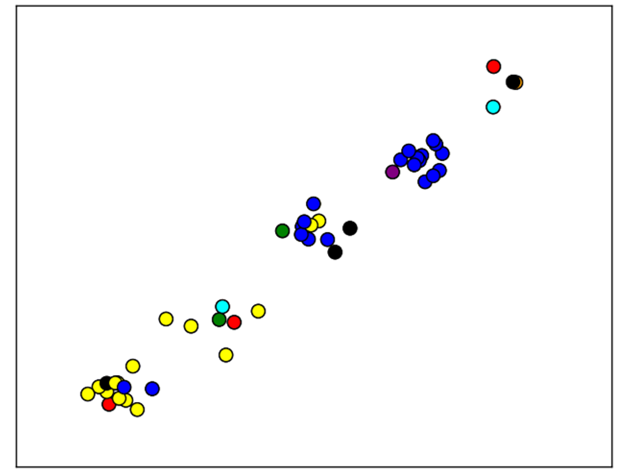}
		\label{fig:avl_wt}
	}
	\caption{Clustering comaprison in the AVL corpus.}
	\label{fig:avl_clust}
\end{figure}

To understand the errors the different techniques are making, Figure~\ref{fig:avl_clust} presents a visualization of the clusterings. On the leftmost side of the figure are the ground truth labels clustering. The coordinates of the points were set up artificially so that segments in the same cluster are close to each other. Also, the correspondence between cluster colors and topics is only meaningful in the true clustering, as the other techniques are agnostic to what topics they discovered.

Intuitively the clustering obtained with the topic model approach seems to be the worst one. Overall, the points that should belong to the same cluster are scattered across different clusters. The only exception is the \ac{BST} topic, but still has the problem of also having been merged with the Balanced Trees topic. The Spectral clustering shows improvements, since the segments of \ac{AVL} Rebalanced have been almost uniquely grouped together. Despite this, it is possible to observe a yellow cluster that contains segments from almost all different topics.

As for as the Walktraps clusterings, we can see that two main clusters were formed (yellow and blue). This means that two topics are being merged: \ac{BST} and Balanced Trees, and \ac{AVL} Rebalance and Tree rotation. These mistakes are probably due to the vocabulary sharing across the topics.

\subsection{Results with the 4S Corpus}\label{sec:ssss_exps}
In this section, we report the document relationship identification experiments done with the \ac{4S} corpus. The experimental setup was, for all algorithms, similar to the one described in Section~\ref{sec:avl_exps}.

\subsubsection{Clustering}
The clustering algorithms results are in Table~\ref{tab:clus_bl_4s}. Spectral clustering obtained again the best performance. The heat map analyzes of the \ac{4S} corpus similarity graph provided similar conclusions to the \ac{AVL} corpus. Therefore, many segments are wrongly perceived as similar by the metrics.
\begin{table}[!ht]
	\centering
	\begin{tabular}{cC{1.2}C{1.2}C{0.9}}
		\toprule
		\textbf{Clustering Algorithm} & $\pmb{ARI}$ & $\pmb{F_1}$ & $\pmb{Acc}$\\
		\midrule
		k-means & 0.04 & 0.33 & 0.37\\ \midrule
		Agglomerative & 0.16 & 0.38  & 0.47 \\ \midrule
		DBSCAN & 0.08 & 0.32 & 0.38 \\ \midrule
		Mean Shift & 0.03 & 0.32 & 0.40 \\ \midrule
		Spectral & \textbf{0.27} & \textbf{0.41} & \textbf{0.47}\\ \midrule
		\ac{NMF} & 0.21 & 0.34 & 0.43 \\
		\bottomrule
	\end{tabular}
	\caption{Results of the clustering algorithms in the \ac{4S} corpus. Best results are in bold.}
	\label{tab:clus_bl_4s}
\end{table}

\subsubsection{Topic Models}
In the topic models experiments, using $score_c$ with the top-$180$ words yield the best results: $ARI = 0.19$, $F_1 = 0.32$, $Acc = 0.45$. Therefore, we establish this parameterization as our topic modeling baseline in the \ac{4S} corpus. Again, changing the scoring function did not impact the results significantly.


\subsubsection{Graph-communities Detection}
The results obtained with the graph-community detection algorithms are in Table~\ref{tab:gcd_4s}\footnote{The results using the Leading Eigenvector were not possible to obtain due to problems in the ARPACK eigen solver.}. In these experiments the Louvain technique performed better in all metrics. It should be noted that the best weighting scheme and scoring function were Count and Best \textit{tf-idf}. These two combinations are different from one another in the sense that one just uses raw counts whereas the other takes into account an importance value of the words. The fact that different scoring functions worked better in different scenarios, shows that the corpora have particular characteristics and, thus, different ways to achieve a better model for the data are necessary.
\begin{table}[!ht]
	\centering
	\begin{tabular}{C{1.5}C{1.4}C{1.5}C{0.6}C{0.6}C{0.6}}
		\toprule
		\textbf{Algorithm} & \textbf{Weighting} & \textbf{Scoring Function} & $\pmb{ARI}$ & $\pmb{F_1}$ & $\pmb{Acc}$\\
		\midrule
		LP & - & $score_{tfif}$ & 0.06 & 0.17 & 0.26\\ \midrule
		\ac{CNM} & - & $score_{seg}$ & 0.06 & 0.25 & 0.35 \\ \midrule
		Louvain & Count & $score_{seg}$ & \textbf{0.21} & \textbf{0.36} & \textbf{0.42} \\ \midrule
		Walktraps & \specialcell{Count + \\ Best tf-idf} & $score_{seg}$ & 0.15 & 0.30 & 0.40\\ \midrule
		Bigclam & - & $score_{seg}$ & 0.05 & 0.09 & 0.19 \\
		\bottomrule
	\end{tabular}
	\caption{Best results obtained in the \ac{4S} corpus. In bold are the highest scores obtained.}
	\label{tab:gcd_4s}
\end{table}

\begin{figure}[!ht]
	\centering
	\includegraphics[scale=0.22]{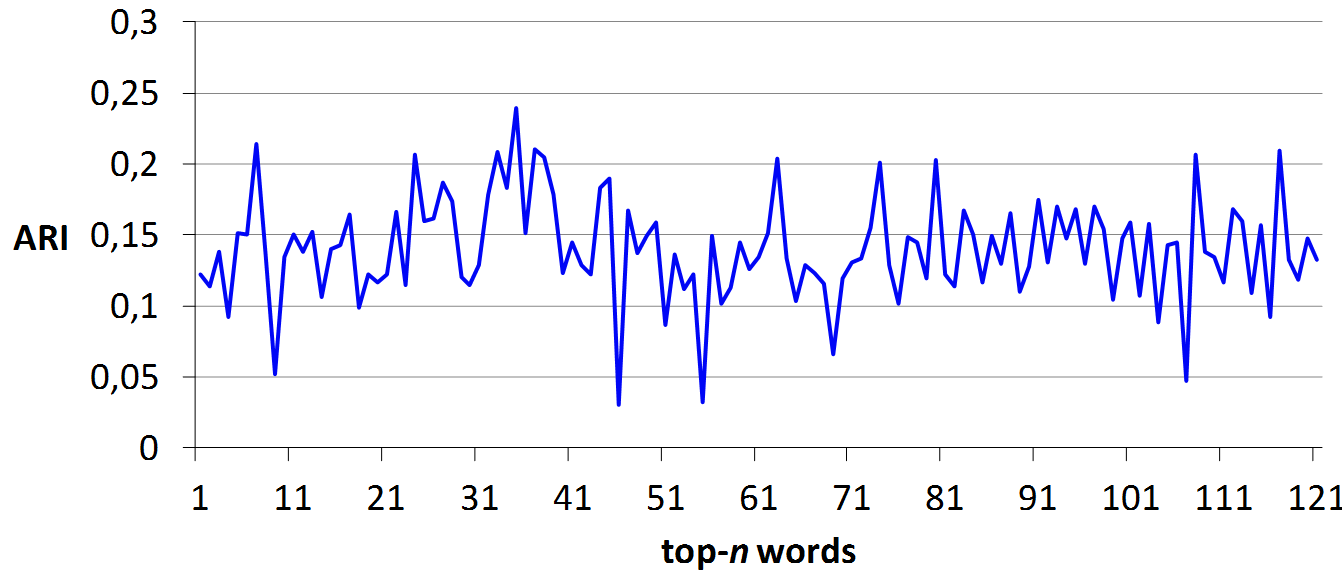}
	\caption{Results obtained in the \ac{4S} corpus by varying the top-$n$ words in the Louvain algorithm.}
	\label{fig:louvain_ssss_topn}
\end{figure}

Given the previous results, we performed the experiments varying the top-$n$ words with the Louvain technique and the Count weighting scheme. The results obtained are in Figure~\ref{fig:louvain_ssss_topn}. As with other similar experiments, oscillations in the results were observed. The best results were obtained with the top-$5$ words with 0.26, 0.40, and 0.48 scores in $ARI$, $F_1$, and $Acc$, respectively.

\subsubsection{Discussion}
After having carried out all experiments with the \ac{4S} corpus, this section provides a discussion of the obtained results. In Table~\ref{tab:res_all_4s}, a summary of all results is provided. Contrary to the \ac{AVL} corpus, results were close between two techniques: Spectral Clustering and Louvain. The former obtained a better overall performance, since it had the highest scores in two of the metrics.
\begin{table}[!ht]
	\centering
	\begin{tabular}{cC{1.5}C{1}C{1}}
		\toprule
		& $\pmb{ARI}$ & $\pmb{F_1}$ & $\pmb{Acc}$\\
		\midrule
		Spectral Clustering & \textbf{0.27} & \textbf{0.41} & 0.47\\ \midrule
		Topic Models & 0.17 & 0.30 & 0.46 \\ \midrule
		Louvain & 0.26 & 0.40 & \textbf{0.48} \\
		\bottomrule
	\end{tabular}
	\caption{Summary of the results in the \ac{4S} corpus. In bold are the highest scores obtained.}
	\label{tab:res_all_4s}
\end{table}

\begin{figure}[!ht]
	\newcommand{\imgwidth}{0.25}
	\centering
	\subfloat[Subfigure 1 list of figures text][True clusters]
	{
		\includegraphics[scale=\imgwidth]{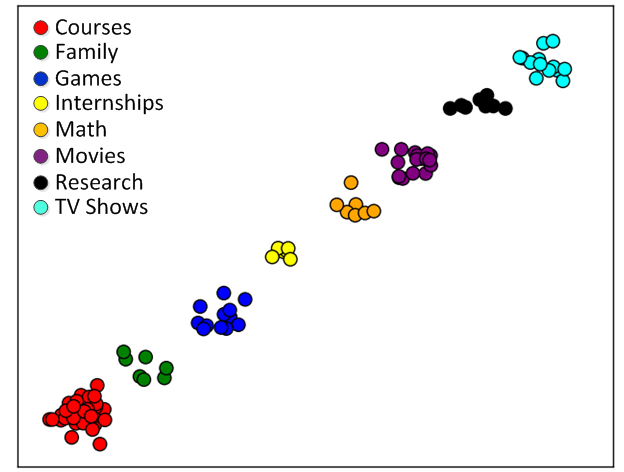}
		\label{fig:ssss_true}
	}
	\subfloat[Subfigure 2 list of figures text][Spectral clustering]
	{
		\includegraphics[scale=\imgwidth]{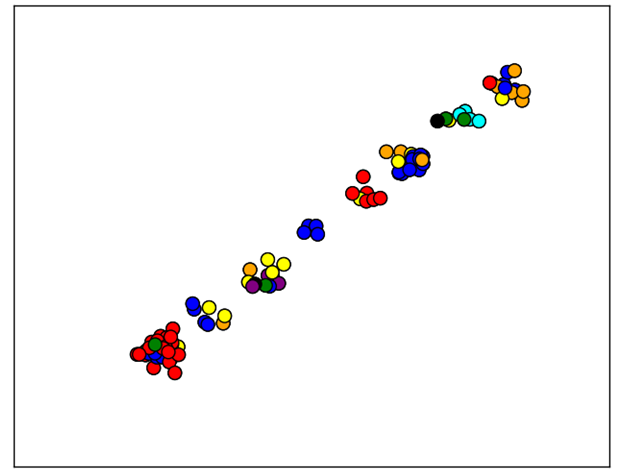}
		\label{fig:ssss_spectral}
	}
	\\
	\subfloat[Subfigure 2 list of figures text][Topic Models clustering]
	{
		\includegraphics[scale=\imgwidth]{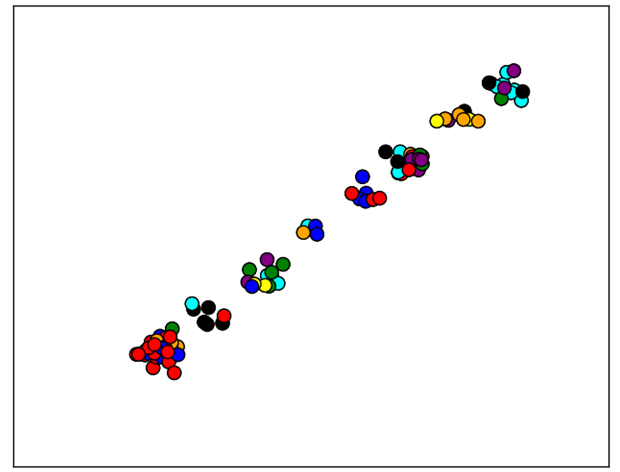}
		\label{fig:ssss_lda}
	}
	\subfloat[Subfigure 2 list of figures text][Louvain clustering]
	{
		\includegraphics[scale=\imgwidth]{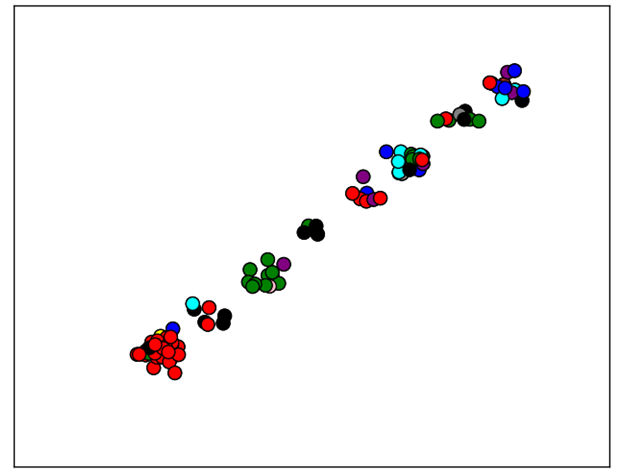}
		\label{fig:ssss_louvain}
	}
	\caption{Clustering comparison in the \ac{4S} corpus.}
	\label{fig:ssss_clust}
\end{figure}

In order to have a visual perception of the types of errors the surveyed techniques are making, Figure~\ref{fig:ssss_clust} shows the corresponding obtained clusterings. From the figure it is possible to conclude that Spectral clustering (Figure~\ref{fig:ssss_spectral}) is the most accurate one. One of the problems is again the merging of topics. Spectral clustering merged the topic of ``Courses'' with ``Math'' (red cluster). This makes sense, since math is a type of course and, thus, the same type of discussions might arise when talking about this topic. The general problem is that the remaining clusters are scattered across all topics. As for as the remaining clusters, they scattered the topics much more. For example, the blue cluster seems to group segments from the ``Movies'' topic, but also includes segments from all other topics. This kind of issue generally occurs in the clusterings obtained with all other techniques.

The advantage of the Louvain clusterings (Figure~\ref{fig:ssss_louvain} over the Topic Models one (Figure~\ref{fig:ssss_lda}) is that the ``Games'' topic was better captured. The problem is that these approaches discover a number of topics higher than the actual value. This reflects how different the \ac{4S} corpus is from the \ac{AVL} corpus. The \ac{4S} corpus contains more heterogeneous segments (even when considering a common topic), whereas the \ac{AVL} segments have an overarching topic.

\section{Conclusion and Future Work}
In this paper we proposed a methodology for discovering cross-document topic segment relationships. Given that this task is done in a collection of related documents, we expected some document segments to be similar and not equivalent, since they have a common overarching topic, but also dissimilar and equivalent (they can come from different media). This dichotomy in syntactic similarity motived us to develop a graph-community detection approach instead of a direct clustering approach. This allows the exploration of a word co-occurrence graph that better models interactions between words in different documents. The discovery of topic segment relationships is not a straightforward process as the output of a graph-community detection task is clusters of words. Therefore, we developed an approach that assigns topic segments to word communities based on a scoring function. Topic segments are considered as equivalent if they were assigned to the same word community.

An evaluation of the proposed approach was made using subsets of a corpus with learning materials in the \ac{AVL} tree domain using the \ac{4S} corpus. The results were not unanimous. In the \ac{AVL} corpus a Walktraps graph-community approach performed better, corroborating our hypothesis that clustering approaches are not suitable for this task. In the \ac{4S} corpus, the Spectral clustering performed better, indicating that segments that more loosely relate with one another at the subtopic level and that do not have a common overarching topic should be modeled differently. It should be noted though that the Louvain technique obtained close results. Therefore, we do not discard the possibility that a graph-community detection approach cannot perform well in this domain. Given these results, a conclusion to draw is that the way documents relate to one another (if they have an overarching topic or not) plays an important role in how their relationships should be discovered.

Regarding future work, one of the main questions that the results leave open is whether the \textit{tf-idf} is a good measure of importance of the words. The argument is that if it was indeed, then straighter correlation in the number of words used and the evaluation metrics should be observed, rather than the constant oscillations. Therefore, we will consider the investigation of other criteria to perform the initial word filtering. Another possibility that can lead to the improvement of the results is the use of edge pruning. Co-occurrence graphs are characterized by high indegree nodes. Therefore, if we prune the edges with less weight, it is possible that more representative word communities are found. The way edges are weighted is another research direction to follow. In this work we weigh all the edges in the graph in the same way. This does not take into account indicators such as the relative position of the words. This can be relevant since topic segments can have a considerable length and, thus, it might be important to consider closer words as having a stronger relation with each other. Finally, we also want to improve the scoring function in order to take into account possible patterns in topic sequences. This means we expect some topics to appear frequently in documents in the same order. This is especially relevant in documents that correspond to learning materials, since many concepts have both a logical and hierarchical structure.

%
\bibliographystyle{ieeetr}

\end{document}